\pdfoutput=1

\documentclass[11pt]{article}

\usepackage{acl}

\usepackage{times}
\usepackage{latexsym}

\usepackage[T1]{fontenc}

\usepackage[utf8]{inputenc}

\usepackage{microtype}

\usepackage{diagbox}
\usepackage{graphicx}
\usepackage{color}
\usepackage{array}
\usepackage{booktabs}
\usepackage{float}
\usepackage{stfloats}
\usepackage{multirow}

\usepackage{arydshln}
\usepackage{natbib}
\title{Prompt Combines Paraphrase: Teaching Pre-trained Models to Understand Rare Biomedical Words
}

\author{
Haochun Wang$^{1}$, Chi Liu$^{1}$, Nuwa Xi$^{1}$, Sendong Zhao$^{1}\thanks{ \ \ Corresponding author}$, Meizhi Ju$^{2}$, Shiwei Zhang$^{2}$\\ \bf{Ziheng Zhang$^{2}$, Yefeng Zheng$^{2}$, Bing Qin$^{1}$ and Ting Liu$^{1}$}
 \\
$^1$Research Center for Social Computing and Information Retrieval, \\Harbin Institute of Technology, China \\
$^2$Tencent Jarvis Lab, Shenzhen, China\\
\small{\texttt{\{hcwang,cliu,nwxi,sdzhao,bqin,tliu\}@ir.hit.edu.cn}} \\
\small{\texttt{meizhi.ju@outlook.com,}} \small{\texttt{\{zswvizhang,zihengzhang,yefengzheng\}@tencent.com}}
}

\begin{document}
\maketitle
\begin{abstract}

Prompt-based fine-tuning for pre-trained models has proven effective for many natural language processing tasks under few-shot settings in general domain. However, tuning with prompt in biomedical domain has not been investigated thoroughly. Biomedical words are often rare in general domain, but quite ubiquitous in biomedical contexts, which dramatically deteriorates the performance of pre-trained models on downstream biomedical applications even after fine-tuning, especially in low-resource scenarios. We propose a simple yet effective approach to helping models learn rare biomedical words during tuning with prompt. Experimental results show that our method can achieve up to 6\% improvement in biomedical natural language inference task without any extra parameters or training steps using few-shot vanilla prompt settings.
\end{abstract}

\section{Introduction}

\begin{figure*}[h]
    \centering
    \includegraphics[width=\textwidth]{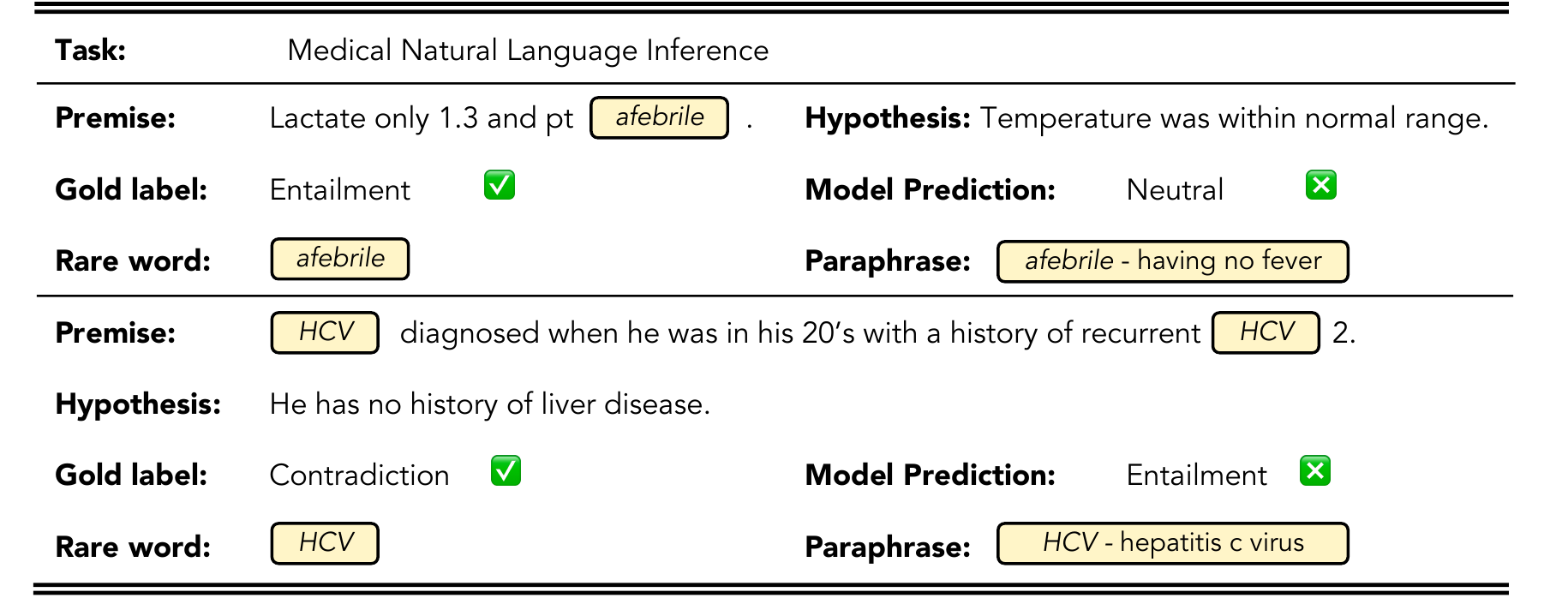}
    \caption{Failures of a biomedical pre-trained model~\citep{lewis2020pretrained} on the biomedical NLI task.}
    \label{wrong-case}
\end{figure*}

Pre-trained models have achieved a great success in natural language processing (NLP) and become a new paradigm for various tasks~\citep{peters2018deep,devlin2019bert,liu2019roberta,qiu2020pre}. Many studies have paid attention to pre-trained models in biomedical NLP tasks~\citep{lee2020biobert,lewis2020pretrained,zhao2021recent}. However, plain pre-trained models sometimes cannot do very well in biomedical NLP tasks. 
In general, there are two challenges to fully exploit the potential of the pre-trained models for biomedical NLP tasks, i.e., (1) \textbf{limited data} and (2) \textbf{rare biomedical words}.
Firstly, it is common that the amount of biomedical labeled data is limited due to strict privacy policy constraints~\citep{vsuster2017short}, high cost and professional requirement for data annotation.
Pre-trained models perform poorly with few samples since abundant training samples are essential to optimize task-related parameters~\citep{liu2021pre}.
Secondly, biomedical words are usually low-frequency words but critical to understanding biomedical texts.
As an example of natural language inference (NLI) task in Figure~\ref{wrong-case}, the model goes wrong during tuning when faced with a rare word ``\textit{afebrile}''\footnote{There are around 4 billion words in the selected biomedical pre-training texts while ``afebrile'' appears only about 100,000 times, accounting for 0.0025\%.}
in the premise, whose meaning is ``\textit{having no fever}''. It can be no easy for the pre-trained models to predict the correct label if the models haven't seen the rare biomedical words for enough times during pre-training or tuning stage. Thus, pre-trained models cannot capture the precise semantics of biomedical texts in the scenario of low-resource tasks.

\begin{figure*}
    \centering
    \includegraphics[width=\textwidth]{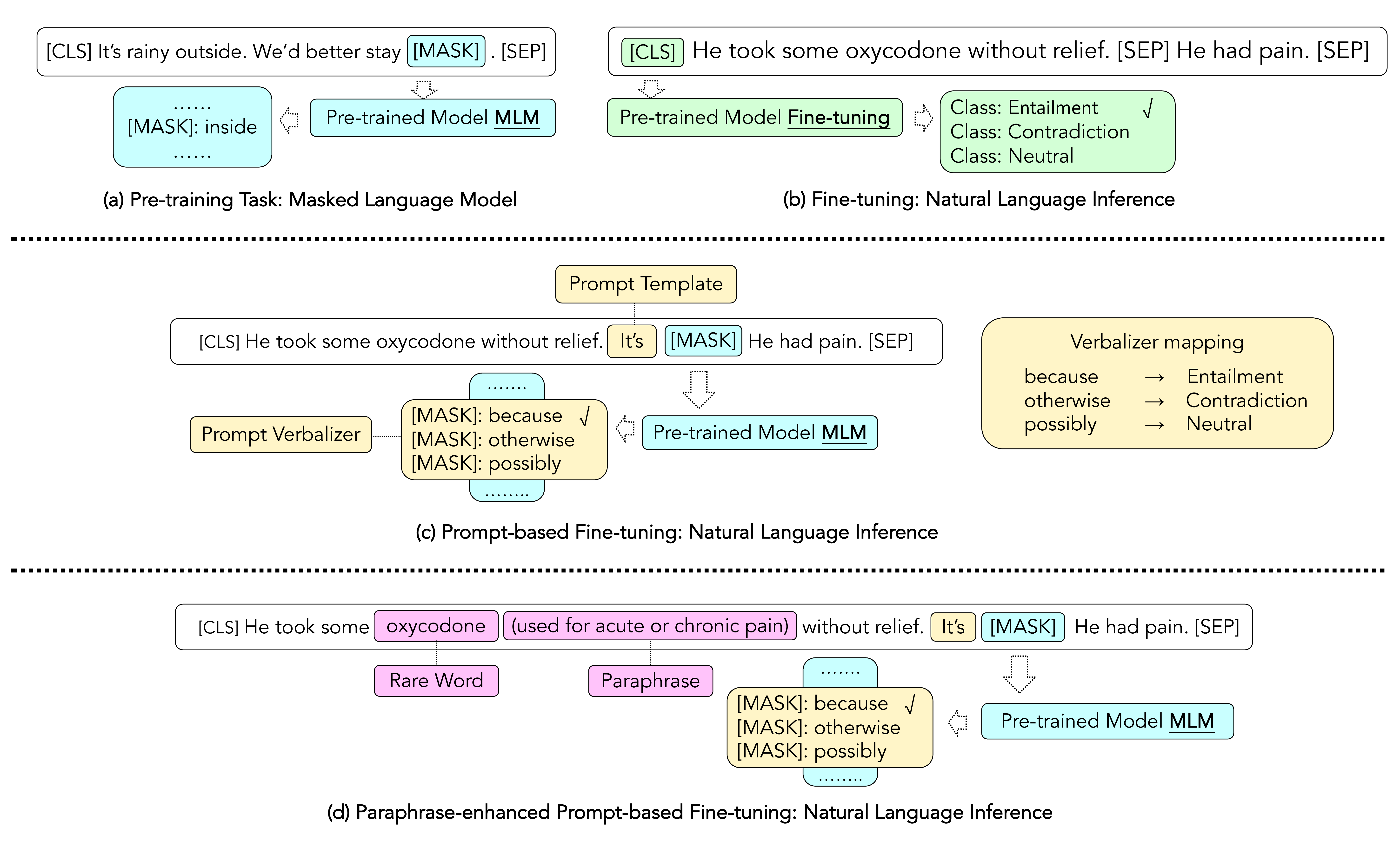}
    \caption{Examples for paradigms of (a) Masked Language Model (MLM) pre-training; (b) Task-specific fine-tuning; (c) Prompt-based fine-tuning, with same task as pre-training process; (d) Paraphrase-enhanced prompt-based fine-tuning. Best viewed in color.}
    \label{intro_prompt}
\end{figure*}

With very few annotated samples available for a new task, it is hard to effectively fine-tune pre-trained models with the additional task-specific parameters, which is even more of a challenge to biomedical domain as mentioned above. Prompt technique has been introduced to smooth the fine-tuning process in the few-shot settings by narrowing down the gap between pre-training stage and the downstream task in general domain~\citep{liu2021pre}, as demonstrated in Figure~\ref{intro_prompt}. Therefore, it is beneficial to adapt prompt-based tuning to biomedical NLP tasks.




Although the challenge of rare words is a critical problem for the biomedical pre-trained models, only a handful of works have studied the issue and most of them focus on enriching the representation of rare words through pre-training stage~\citep{schick2020rare,yu2021dict,wu2020taking}.
Thus, it naturally requires them to involve a second-round pre-training or further training steps with biomedical knowledge to achieve the above goal, which is highly time-consuming and inefficient. Alternatively, we emphasize on tuning stage instead of pre-training to resolve these issues. When coming across an unknown word, human may seek the dictionary for its paraphrase. Enlightened by this phenomenon, we propose to explain rare biomedical words with the paraphrases on the basis of prompt-based tuning. The new approach could enhance tuning capability in understanding biomedical words. Furthermore, as a generic plug-in module for non-specific datasets, our approach is model-agnostic and can be easily transferred to other domains.\footnote{We release our code at \url{https://github.com/s65b40/prompt\_n\_paraphrase}}




In summary, our contributions are as follows:
\begin{itemize}
    \item We investigate a valuable problem of the adaptation of pre-trained models to few-shot scenarios to enhance biomedical text understanding with a focus on rare biomedical words.
    \item We propose a novel method to combine the prompt paradigm and paraphrases of rare biomedical words in the tuning stage of pre-trained models to address the limitation caused by ``rare but key words'' in biomedical texts.
    \item We evaluate on six pre-trained models over two biomedical natural language understanding datasets$-$MedNLI and MedSTS. Our approach can improve the performance by up to 6\% in the few-shot settings. Moreover, we discuss how the paraphrases improve the pre-trained models and provide a perspective about task-related rare words.
\end{itemize}


\section{Related Work}
\paragraph{Representation learning of rare words in pre-trained models.}
Words in the vocabulary list follow a Zipf distribution~\citep{zipf2016human} by and large. Previous works have shown that the word representation space of the pre-trained models is anisotropic and high-frequency words dominate the representation of a sentence which can induce semantic bias~\citep{gao2018representation,li-etal-2020-sentence,yan-etal-2021-consert}. Meanwhile, it has also been proven that rare words limit the performance of pre-trained models as the rare words can play a decisive role in the sentence understanding~\citep{schick2020rare,wu2020taking,yu2021dict}. \citet{schick2020rare} introduced one-token approximation to infer the embedding of arbitrary rare word by a single token. \citet{wu2020taking} proposed to take notes on the fly to maintain a note dictionary for rare words to save the contextual information that helps enhance the representation during pre-training.

\paragraph{Biomedical pre-trained models.}
With the booming trend of pre-trained models in NLP tasks~\citep{peters2018deep,devlin2019bert,liu2019roberta}, various trials have been made to investigate the pre-trained models in biomedical domain~\citep{peng2019transfer,lee2020biobert,huang2019clinicalbert}. 
\citet{lewis2020pretrained} and \citet{gu2021domain} further built the domain-specific vocabulary to amend the representation of biomedical words. More recent works guided biomedical pre-trained models with domain knowledge. For example, \citet{zhang2021smedbert} amplified the biomedical entities with type information from neighbor entities. \citet{michalopoulos2021umlsbert} learned clinical word embeddings with the association of synonyms in the Unified Medical Language System (UMLS) Metathesaurus. 

\paragraph{Tuning pre-trained models with prompt.}
Many works were dedicated to applying prompt in fine-tuning by adapting the downstream tasks to the paradigm of pre-training tasks. Prompts that have been employed by now can be categorized into two groups: (1) discrete prompt in natural language~\citep{schick2021exploiting, gao-etal-2021-making} and (2) continuous prompt in representation based on trainable vectors~\citep{li-liang-2021-prefix,shin2020eliciting}. Discrete prompt follows the settings of pre-training tasks and converts the downstream tasks into a cloze question format without requiring additional parameters. Continuous prompt inserts prompt embeddings into the models, which could perform better than discrete prompt but at the expense of explainability and extra training cost on additional data for the prompt embeddings~\citep{wei2021finetuned, gu2021ppt}. Recently, \citet{liu2021p} considered training continuous prompts as a parameter-efficient method instead of tuning the parameters of the entire pre-trained model. In this study, we follow the paradigm of discrete prompt to avoid introducing more ambiguity from the prompt embeddings or training costs on additional training data.
\section{Method}
In this section, we introduce how we find the rare biomedical words and append paraphrases to the rare biomedical words with the prompt-based tuning of pre-trained models in a model-agnostic plug-in manner. 

\subsection{Rare Words}
The rarity of a word mostly depends on its frequency in a certain corpus, which can vary from context to context. A rare word in the pre-training corpora is possibly not \textit{that} rare in the down-stream tasks. In this work, we define the ``rare words'' as the words whose frequency is under a specific threshold in the pre-training corpora as aforementioned. 

Meanwhile, although the pre-trained models tokenize the input words into tokens, tokenizers based on byte-pair encoding~\citep{sennrich2016neural} or WordPiece~\citep{6289079} split words into sub-words by frequency or likelihood, which is both dominated by the common words. Thus, although the rare words can be split into possible non-rare tokens, there is not much semantics from the original rare words retained after being tokenized into common tokens for the pre-trained models. Also, tokenizers of different pre-trained models can tokenize the same rare word into different tokens and consequentially make rare tokens model-related. For example, BERT-Large~\citep{devlin2019bert} model tokenizes ``\textit{afebrile}'' into ``\textit{af-eb-ril-e}'' while Biomedical-Clinical-RoBERTa-Large~\citep{lewis2020pretrained} model tokenizes it into ``\textit{a-fe-brile}''.

\subsection{Selection of Rare Biomedical Words}
\label{bc-roberta}
To obtain the frequency of words, we adopt the biomedical corpora including PubMed abstract,\footnote{\url{https://pubmed.ncbi.nlm.nih.gov}} PubMed Central\footnote{\url{https://www.ncbi.nlm.nih.gov/pmc}} (PMC) full-text and MIMIC-III dataset,\footnote{\url{https://physionet.org/content/mimiciii/1.4/}} which are widely used for pre-training biomedical language models, such as BC-RoBETRa~\citep{lewis2020pretrained}, BioBERT~\cite{lee2020biobert}, and PuBMedBERT~\cite{pubmedbert}. We loop the above corpora to obtain the frequency of each word in the pre-training phase. The rare words found in biomedical corpora are likely to contain words not only in the biomedical domain but also in the general domain. Instead of including all rare words, we consider rare words from biomedical domain with the following two reasons:
(1) \textbf{Domain-specific distribution}: unlike the general domain,  distribution of words in the biomedical domain is shaped with domain-specific terms, such as disease, medicine, diagnosis and treatment~\citep{lee2020biobert}. 
(2) \textbf{Task-specific words}: rare words from the biomedical domain can contribute more to biomedical tasks than that from the general domain. Therefore, we introduce a threshold, an empirical hyper-parameter, to assist the selection of rare words following~\citet{yu2021dict}, and we also experiment with different thresholds for the rare words in Section~\ref{threshold}. We retrieve the paraphrases of the rare biomedical words from an online dictionary ``Wiktionary''.\footnote{Wiktionary - \url{https://en.wiktionary.org/}} To optimize the selection, we only keep the rare words that are tagged with medical-related categories from the Wiktionary, i.e. \textit{medical}, \textit{medicine}, \textit{disease}, \textit{symptom} and \textit{pharmacology}.


\subsection{Selection of Paraphrases}
\label{selection_paraphrase}
There can be more than just one paraphrase for a rare biomedical word and it is tricky to choose the most appropriate paraphrases. Therefore, to avoid introducing noise from the inappropriate paraphrases, we exclude rare biomedical words with more than one corresponding paraphrase. In addition, we ignore paraphrases that contain additional rare words whose frequencies are below the set threshold since it only replaces one rare word with another. Meanwhile, considering that biomedical abbreviations are likely to be tokenized into separate letters with no meaningful semantic information, we retrieve and append the paraphrases to all the biomedical abbreviations. 

\subsection{Prompt-based Fine-Tuning with Paraphrases}
When coming across new words during reading, humans habitually seek dictionaries for the corresponding paraphrases to help us understand. Following the same idea, we guide the pre-trained models with paraphrases, where rare words are followed by the parenthesis punctuation, as shown in Figure~\ref{intro_prompt}(d). In this way, given a pre-trained model, paraphrases of biomedical rare words can be considered as a portable plug-in module and generated for any dataset instantly before prompt-based fine-tuning. 
\section{Experiments}

\subsection{Setup}
\paragraph{Models.} 
To demonstrate that our approach is model-agnostic, we adopt six pre-trained models in both general and biomedical domains, namely (1) BERT-Large~\citep{devlin2019bert}, (2) RoBERTa-Large~\citep{liu2019roberta}, (3) BioBERT-Base~\citep{lee2020biobert}, (4) PubMedBERT-Base~\citep{pubmedbert}, (5) SciBERT-Base~\citep{beltagy-etal-2019-scibert} and (6) BC-RoBERTa-Large (Biomedical-Clinical RoBERTa-Large)~\citep{lewis2020pretrained}. During prompt-based tuning, we use the same set of hyper-parameters for all the six pre-trained models, including learning rate of $1 \times 10^{-5}$, batch size of 2 and max epoch of 10.

\paragraph{Datasets.}  Previous work demonstrates that rare words have more impact on Natural Language Understanding (NLU) tasks than Information Extraction (IE)~\citep{schick2020rare}, while most biomedical NLP tasks fall into the category of IE~\citep{shin2020bio,gu2021domain}. To better demonstrate the method effectiveness, we perform evaluation over the two biomedical NLU datasets, namely MedNLI~\citep{romanov2018lessons} and MedSTS\footnote{We use ClinicalSTS-2018 which is a sub-dataset of MedSTS provided by the maintainers of the MedSTS project.}~\citep{wang2020medsts}.


MedNLI is a natural language inference dataset where premises are selected from real clinical notes in MIMIC-III~\citep{johnson2016mimic}. And MedSTS is a semantic textual similarity dataset gathered from a clinical corpus at Mayo Clinic and the ground-truth label of the similarity is the mean of the subjectively annotated scores from multiple annotators. As MedSTS is actually a regression task, we adapt the task following~\citet{gao-etal-2021-making} and convert it into a classification task. We use the same data splitting of training, development and test sets as the original two datasets. Statistics of datasets can be found in Table~\ref{data_stat}.  

\begingroup
\setlength{\tabcolsep}{8pt} 
\renewcommand{\arraystretch}{1.2} 
\begin{table}[H]
\centering
\scalebox{0.7}{
\resizebox{\columnwidth}{!}{%
\begin{tabular}{l|ccc}
\hline
\hline
\textbf{Dataset}                  & \textbf{Train} & \textbf{Dev}  & \textbf{Test} \\ \hline
MedNLI & 11,232 & 1,395 & 1,422 \\
\hline
MedSTS & 750   & /    & 318  \\ \hline \hline
\end{tabular}%
}}
\caption{Statistics of the MedNLI and MedSTS datasets. We use the ClinicalSTS-2018 subset of MedSTS.}
\label{data_stat}
\end{table}
\endgroup

\paragraph{Few-shot datasets.} 
Initialized with 10 different random seeds, we randomly sample instances within the range of 16 to 256 from corresponding training and development sets as the few-shot training and development sets. The original test set is directly used as the few-shot test set. Note that there is no development set in MedSTS, so we sample the few-shot development set from the original training set with the same quantity of samples as the few-shot training set with no overlapping instances. Accuracy and Pearson correlation coefficients are used as the evaluation metrics for MedNLI and MedSTS, respectively.

\paragraph{Prompt settings.} 
We prepare discrete prompts using the same prompt settings from~\citet{schick2021exploiting} and~\citet{gao-etal-2021-making}, which correspond to the NLI and STS tasks respectively in Table~\ref{prompt}.

\begin{table}[H]
\renewcommand{\arraystretch}{1.2}
\resizebox{\columnwidth}{!}{%
\begin{tabular}{c|c|c}
\hline
\hline
\textbf{Task} & \textbf{Template}                                          & \textbf{Verbalizer} \\
\hline
MedNLI        & \textless{}Sent1\textgreater{}. {[}MASK{]}. \textless{}Sent2\textgreater{} & Yes/No/maybe           \\
MedSTS        & \textless{}Sent1\textgreater{}. {[}MASK{]}. \textless{}Sent2\textgreater{} & Yes/No    \\
\hline
\hline
\end{tabular}%
}
\caption{Prompt settings for MedNLI and MedSTS.}
\label{prompt}
\end{table}

\paragraph{Rare biomedical words and paraphrases.}
We find that a threshold of rare biomedical words at 200,000, corresponding to a frequency less than 0.005\% in the pre-training corpora, can yield better results in most scenarios 
(details in Section~\ref{threshold}), so we consider the biomedical words that appear less than 200,000 times as ``rare biomedical words'' and prepare the same rare word and paraphrase sets for all the models to validate the generalization ability of our approach. 

\subsection{Few-shot learning results}
We report the mean accuracy for the MedNLI task and Pearson correlation coefficient for the MedSTS task over 10 sampled few-shot datasets based on different random seeds, along with standard deviation and \textit{p}-value of paired \textit{t}-test. Table~\ref{few-shot-task-mednli} and Table~\ref{few-shot-task-sts} show the results for the two tasks on six pre-trained models.

\paragraph{Results on MedNLI.} 
The pre-trained models with the paraphrases for rare biomedical words can outperform the baselines in all cases and can bring about 6\% improvement on average for few-shot learning with 16 training samples and 2\% with 256 training samples. All performance improvements are statistically significant with \textit{p}-value less than 0.05 except only two out of ten cases from the RoBERTa-based model in general domain. 
Besides, small pre-trained biomedical models are comparable with large pre-trained models in the general domain under the few-shot settings. Furthermore, with more training samples up to 256, our approach is consistently effective.

\paragraph{Results on MedSTS.} 
Similar to MedNLI, the incorporation of paraphrases improves the performances compared with baselines in general. For some cases, statistical significance is not as stable as that on MedNLI for two reasons: 
 (1) Some ground-truth labels in the MedSTS task can be biased due to the subjectivity of annotation~\citep{yang2020measurement}; 
 (2) Rare biomedical words shared in the sentence pair of the same sample can be a shortcut for the pre-trained models, as the paraphrases increase the overlap between the two sentences and mislead the models to overlook the rest of the sentences~\citep{mccoy2019right}.

\begin{table*}[h]
\centering
\scalebox{0.8}{

\resizebox{\textwidth}{!}{
\begin{tabular}{r|ccccc}
\hline
\hline
\multicolumn{6}{c}{MedNLI} \\

\hline
\diagbox{Model}{\#Samples} & 16 & 32 & 64 & 128 & 256\\
\hline
\multicolumn{1}{l|} {BERT-Large}  & 38.9 (3.7) & 44.5 (5.2) & 50.1 (5.2) & 54.8 (2.5) & 59.9 (1.2) \\
+ paraphrase & \textbf{40.8} (4.1) & \textbf{46.0} (5.5) & \textbf{53.3} (4.9) & \textbf{58.1} (1.4) & \textbf{61.9} (1.4) \\
\cdashline{2-6}
\textit{p}-value     & < 0.01 & 0.02 & < 0.01 & < 0.01 & < 0.01 \\

\hline

\multicolumn{1}{l|} {RoBERTa-Large} & 43.2 (6.7) & 52.1 (8.2) & 63.6 (4.6) & 69.2 (1.8) & 72.7 (1.4) \\
+ paraphrase & \textbf{49.5} (8.1) & \textbf{56.1} (7.6) & \textbf{65.6} (2.9) & \textbf{70.8} (0.7) & \textbf{74.0} (1.3) \\
\cdashline{2-6}
\textit{p}-value     & < 0.01 & 0.03 & 0.08 & 0.06 & 0.02 \\
\hline

\multicolumn{1}{l|} {BioBERT-Base} & 34.1 (1.5) & 38.5 (3.3) & 42.5 (4.6) & 52.1 (2.5) & 59.4 (1.6) \\
+ paraphrase & \textbf{36.3} (1.8) & \textbf{40.9} (3.5) & \textbf{45.2} (4.3) & \textbf{54.0} (2.3) & \textbf{60.4} (4.7) \\
\cdashline{2-6}
\textit{p}-value     & < 0.01 & < 0.01 & < 0.01 & 0.02 & 0.03 \\
\hline

\multicolumn{1}{l|} {PubMedBERT-Base} & 40.5 (3.4) & 46.8 (4.9) & 53.9 (4.0) & 62.9 (1.6) & 69.2 (1.1) \\
+ paraphrase & \textbf{45.0} (4.0) & \textbf{49.7} (5.0) & \textbf{56.4} (3.4) & \textbf{65.8} (1.6) & \textbf{71.0} (1.6) \\
\cdashline{2-6}
\textit{p}-value     & < 0.01 & < 0.01 & < 0.01 & < 0.01 & < 0.01 \\
\hline

\multicolumn{1}{l|} {SciBERT-Base} & 36.8 (1.2) & 41.4 (3.3) & 49.0 (3.7) & 54.7 (2.0) & 60.9 (1.0) \\
+ paraphrase & \textbf{38.2} (2.1) & \textbf{45.4} (5.2) & \textbf{50.1} (3.5) & \textbf{56.4} (2.2) & \textbf{61.9} (1.6) \\
\cdashline{2-6}
\textit{p}-value     & < 0.01 & 0.03 & 0.01 & 0.02 & < 0.01 \\
\hline

\multicolumn{1}{l|}{BC-RoBERTa-Large} & 51.3 (5.9) & 60.6 (6.7) & 71.0 (3.7) & 80.6 (1.3) & 83.1 (1.3) \\
+ paraphrase & \textbf{56.6} (5.0) & \textbf{62.3} (6.0) & \textbf{74.5} (3.0) & \textbf{81.1} (1.5) & \textbf{83.6} (1.0) \\
\cdashline{2-6}
\textit{p}-value     & < 0.01 & 0.05 & < 0.01 & 0.02 & 0.01 \\

\hline
\hline
\end{tabular}}}
\caption{
Few-shot results on the MedNLI task using various pre-trained models with training and development sets of different sizes. We report mean (standard deviation) performance of accuracy over 10 different random seeds, along with the \textit{p}-value of the paired \textit{t}-test. + paraphrase: with paraphrases of selected rare biomedical words.}
\label{few-shot-task-mednli}

\end{table*}

\begin{table*}[h!]
\renewcommand{\arraystretch}{1.05}
\centering
\scalebox{0.8}{

\resizebox{\textwidth}{!}{
\begin{tabular}{r|ccccc}
\hline
\hline
\multicolumn{6}{c}{MedSTS} \\
\hline
\diagbox{Model}{\#Samples} & 16 & 32 & 64 & 128 & 256\\
\hline
\multicolumn{1}{l|} {BERT-Large} & 14.1 (7.4) & 24.8 (10.1) & 43.6 (5.7) & 60.2 (4.7) & 72.2 (4.5) \\
+ paraphrase & \textbf{18.5} (9.0) & \textbf{28.2} (12.1) & \textbf{48.7} (7.1) & \textbf{64.1} (5.1) & \textbf{72.7} (4.6) \\
\cdashline{2-6}
\textit{p}-value     & < 0.01 & 0.04 & < 0.01 & 0.01 & 0.08 \\
\hline

\multicolumn{1}{l|} {RoBERTa-Large} & 29.5 (9.3) & 41.7 (18.5) & 55.1 (12.1) & 67.6 (6.8) & 76.0 (3.5) \\
+ paraphrase & \textbf{34.6} (13.5) & \textbf{46.1} (13.0) & \textbf{57.7} (12.7) & \textbf{69.9} (6.7) & \textbf{77.2} (3.7) \\
\cdashline{2-6}
\textit{p}-value     & 0.04 & 0.01 & 0.02 & 0.03 & 0.02\\
\hline

\multicolumn{1}{l|} {BioBERT-Base} & 17.3 (14.4) & 26.3 (13.7) & 41.4 (9.0) & 52.2 (10.6) & 63.0 (7.3) \\
+ paraphrase & \textbf{20.0} (12.9) & \textbf{28.2} (12.8) & \textbf{43.1} (8.6) & \textbf{53.7} (9.2) & \textbf{64.2} (7.1) \\
\cdashline{2-6}
\textit{p}-value     & 0.03 & 0.02 & 0.01 & 0.02 & 0.02 \\
\hline

\multicolumn{1}{l|} {PubMedBERT-Base} & 10.3 (9.8) & 22.8 (10.7) & 36.9 (10.6) & 48.1 (11.9) & \textbf{65.5} (9.5) \\
+ paraphrase & \textbf{18.8} (13.7) & \textbf{27.3} (11.6) & \textbf{39.9} (10.4) & \textbf{49.4} (12.4) & 65.1 (9.4) \\
\cdashline{2-6}
\textit{p}-value     & 0.04 & 0.01 & < 0.01 & 0.07 & 0.1 \\
\hline

\multicolumn{1}{l|} {SciBERT-Base} & 15.2 (16.9) & 29.1 (18.2) & 45.5 (14.0) & 60.8 (9.0) & 72.2 (7.3) \\
+ paraphrase & \textbf{19.2} (15.5) & \textbf{32.6} (18.4) & \textbf{48.9} (14.4) & \textbf{62.5} (8.3) & \textbf{73.4} (6.8) \\
\cdashline{2-6}
\textit{p}-value     & 0.01 & < 0.01 & 0.02 & < 0.01 & < 0.01 \\
\hline
\multicolumn{1}{l|} {BC-RoBERTa-Large} & \textbf{54.2} (8.1) & 63.9 (9.2) & 73.3 (3.8) & 77.4 (2.7) & 81.5 (1.5) \\
+ paraphrase & 53.0 (7.4) & \textbf{67.2} (6.6) & \textbf{74.5} (2.7) & \textbf{79.1} (1.6) & \textbf{81.8} (1.2) \\
\cdashline{2-6}
\textit{p}-value     & 0.08 & 0.01 & 0.03 & < 0.01 & 0.02 \\
\hline
\hline
\end{tabular}}}
\caption{
Few-shot results on the MedSTS task using various pre-trained models with training and development sets of different sizes. We report mean (standard deviation) performance of Pearson correlation coefficient over 10 different random seeds, along with the \textit{p}-value of the paired \textit{t}-test. + paraphrase: with paraphrases of selected rare biomedical words.}
\label{few-shot-task-sts}

\end{table*}

\subsection{Thresholds for rare biomedical words}
\label{threshold}
The number of paraphrases of rare biomedical words involved in the samples can directly affect the model performance, which is controlled by the pre-set threshold.
To measure the influence of different thresholds, we conduct experiments over MedNLI task since its labelled data is more objective compared to MedSTS. 

\paragraph{Statistics of rare biomedical words.}
Table~\ref{number_of_rare_word} shows the mean of rare biomedical words in the training, development and test sets under different thresholds for the MedNLI task. 
Among the above thresholds, the number of rare biomedical words varies from 29 to 41 in the test set, while the total number of all rare biomedical words varies from 129 to 196. Each rare biomedical word appears around 5 times on average in the test set.  
Note that as the threshold increases, the number of paraphrases does not necessarily increase as fast as the number of rare biomedical words which is because that although higher threshold does include more rare biomedical words, not all are appended with paraphrases as some of them are excluded as mentioned in Section~\ref{selection_paraphrase}.
\begin{table*}[tb]
\centering
\resizebox{\textwidth}{!}
{%
\begin{tabular}{cc|cccccccccc|c}
\hline
\hline
\multicolumn{2}{c|}{\multirow{2}{*}{\diagbox{threshold}{Dataset}}}&  \multicolumn{2}{c}{16}  & \multicolumn{2}{c}{32} & \multicolumn{2}{c}{64} & \multicolumn{2}{c}{128} & \multicolumn{2}{c|}{256} & \multicolumn{1}{c}{\multirow{2}{*}{Test}}\\
\cline{3-12}
\multicolumn{2}{c|}{\multirow{2}{*}{}}& Train & Dev & Train & Dev & Train & Dev & Train & Dev &  Train & Dev & \multicolumn{1}{c}{\multirow{2}{*}{}} \\
\hline
\multicolumn{1}{c|}{\multirow{2}{*}{\textit{t} = 20k}}& *&1.1  & 1.1  & 1.9  & 1.7   & 3.4  & 3.9  & 6.8  & 7.6  & 13.2 & 14.3 & 29  \\
\cline{2-13}
\multicolumn{1}{c|}{}& **& 1.1 &1.1 &1.9 &1.8  &3.7 &4.4 &7.3 &9.0 &16.3 &21.1 &129 \\
\hline

\multicolumn{1}{c|}{\multirow{2}{*}{\textit{t} = 50k}}&* &1.8 &1.5  &3.5 &2.2  &6.8&4.8  &12.4 &9.2 &21.8&17.0 & 40\\
\cline{2-13}
\multicolumn{1}{c|}{} & ** &1.8&1.5 &3.5&2.3 &7.1&5.4 &13.1&11.2 &26.6&24.9& 196\\
\hline

\multicolumn{1}{c|}{\multirow{2}{*}{\textit{t} = 100k}}&* &1.8&1.6&3.8&2.6&7.2&5.9&14.0&11.1&25.3&20.2&41\\
\cline{2-13}
\multicolumn{1}{c|}{} & ** &1.8&1.6&3.8&2.8&7.6&6.5&14.8&13.0&30.8&25.4&179\\
\hline
\multicolumn{1}{c|}{\multirow{2}{*}{\textit{t} = 200k}}&* &1.9&2.1&4.3&3.5&7.6&6.8&13.9&12.3&23.3&21.7&41\\
\cline{2-13}
\multicolumn{1}{c|}{} & ** &1.9&2.1&4.3&3.7&8.2&7.8&15.3&15.5&33.4&32.6&185\\

\hline
\hline
\end{tabular}
}
\caption{The number of selected rare biomedical words in the training, development and test sets for different few-shot datasets on MedNLI.  ``\textit{t} = 20k'' means the threshold for rare biomedical words is 20k. ``*'' is the number of different rare biomedical words within the threshold. ``**'' denotes the total occurrences of rare biomedical words in the dataset (a rare biomedical word can appear more than once).}
\label{number_of_rare_word}
\end{table*}

In addition to the fixed threshold of 200,000, we experiment with the thresholds ranging from 20,000 to 200,000. Table~\ref{table-threshold} further demonstrates the effectiveness of paraphrases of rare biomedical words in most cases, where 200,000 of the threshold performs the best.\footnote{We also attempt with thresholds higher than 200,000 but it will not bring improvement as much as 200,000.} 

\begin{table}[htbp]
\resizebox{\columnwidth}{!}
{%
\begin{tabular}{c|ccccc}
\hline
\hline
\#Samples& 16 & 32 & 64 & 128 & 256 \\
\hline
\textit{w/o} paraphrase       & 51.3 & 60.6 & 71.0 & 80.6 & 83.1 \\
\textit{t} = 20k & 55.0 & 61.9 & 72.4 & 80.6 & 83.4 \\
\textit{t} = 50k & 54.2 & 62.1 & 73.8 & 81.0 & \textbf{83.7} \\
\textit{t} = 100k & 54.5 & \textbf{62.4} & 73.4 & 80.4 & 83.2 \\
\textit{t} = 200k & \textbf{56.6} & 62.3 & \textbf{74.5} & \textbf{81.1} & 83.6 \\

\hline
\hline
\end{tabular}%
}
\caption{Results for few-shot learning with different thresholds of rare biomedical words on MedNLI with the BC-RoBERTa-Large model. \textit{w/o} paraphrase: without paraphrases for rare biomedical words. \textit{t}: the threshold for rare biomedical words.}
\label{table-threshold}
\end{table}


\section{Discussion}



\paragraph{Train with more samples.} 
Apart from applying paraphrases in few-shot scenarios, we also attempt with more training samples for the MedNLI task, even with full-size training dataset. We attempt to sample the same number of development samples from the official MedNLI development set (with 1,395 samples) to match the number of training samples. If the training set is larger than the full-size development set, we just use the whole development set.

Table~\ref{mednli-more} demonstrates that our method outperforms the baseline on four out of six cases and performs comparably on the remaining two cases. When the whole training set is used, our model with paraphrases achieves 0.8\% improvement.
The minor improvement might be attributed to the ``not so rare'' biomedical words as the rarity of words decreases during the expansion of training set, which helps the pre-trained models learn the semantics of the rare biomedical words better even without paraphrases.

\begin{table*}[htbp]
\centering
\scalebox{0.9}{
\begin{tabular}{r|cccccc}
\hline
\hline
\multicolumn{7}{c}{MedNLI} \\
\hline 
\diagbox{Model}{\#Training} & 512 & 1024 & 2048 & 4096 & 8192 & full-size \\
\hline
BC-RoBERTa-Large & 84.8 (0.7) & 85.5 (0.8) & 86.4 (0.5)&86.3 (0.7)& 86.2 (0.6) & 85.9 (0.6)\\
+ paraphrases & \textbf{85.2} (0.8) &\textbf{86.3} (1.0) & 86.4 (0.7)& 86.3 (0.6)&\textbf{86.7} (0.5) & \textbf{86.7} (0.7)\\
\cdashline{2-7}
\textit{p}-value     & 0.05 & < 0.01 & 0.43 & 0.36 & < 0.01 & 0.02 \\
\hline
\hline
\end{tabular}}
\caption{
Test results on the MedNLI dataset with larger size of training sets. We report mean (and standard deviation) accuracy. + paraphrase: with paraphrases of rare biomedical words.
}
\label{mednli-more}
\end{table*}

\paragraph{Which to look up?} 
Paraphrases can be helpful in general. However, for individual samples, that is not always the case. In Table~\ref{case-study}, we further scrutinize the cases where the model yields different predictions after adding the paraphrases. It is observed that the paraphrases can be less effective if the paraphrases of rare words are not task-related and critical for sentence understanding. For instance, in the third case in Table~\ref{case-study}, ``\textit{CHF}'' is a rare biomedical word which means ``\textit{congestive heart failure}''. Pre-trained models can easily match the hypothesis of ``\textit{poorly functioning}'' with the given premise. Otherwise, paraphrases can possibly introduce confusion. For example, in the wrong case in Table~\ref{case-study}, although ``\textit{workup}'' is a rare biomedical word in the pre-training corpora, it does not severely affect the semantics of the sentence and the pre-trained model turns to make wrong prediction with the misdirection from supplementary paraphrase.  

Actually, humans tend to continue reading unless the unknown words hinder the understanding. With this motivation, we believe that it is worthwhile to explore how to attach \textit{informative} paraphrases for the rare words, which will be investigated in future.

\begin{table*}[hptb]
\renewcommand{\arraystretch}{1.13}
\resizebox{\textwidth}{!}{%
\begin{tabular}{l|c|c}
\hline
\hline
\multicolumn{1}{c|}{Sentence Pairs} & w/o paraphrases & w/ paraphrases \\ \midrule
\begin{tabular}[c]{@{}l@{}}P: She was found to have \textbf{BRBPR} \textit{(bright red blood per rectum)} on rectal exam. \\ H: the patient had bright red blood per rectum\end{tabular}     & Neutral  & \begin{tabular}[c]{@{}c@{}}Entailment\\ (\textbf{right answer})\end{tabular}    \\
\hline
\begin{tabular}[c]{@{}l@{}}P: Antenatal history - pregnancy complicated by chronic hypertension with increased \\ gestational hypertension leading to admission 3 days prior to delivery followed by cesarean\\ section.\\ H: The patient had \textbf{proteinuria} \textit{(The presence of protein in the urine)} during pregnancy\end{tabular} & Entailment      & \begin{tabular}[c]{@{}c@{}}Neutral\\ (\textbf{right answer})\end{tabular}        \\
\hline
\begin{tabular}[c]{@{}l@{}}P: Following this rehab admission she was sent to a different OSH on {[}**2725-10-26**{]}, \\ for acute \textbf{CHF} \textit{(congestive heart failure)} and at least one PEA arrest.\\ H: The patient has a poorly functioning heart.\end{tabular}                                                                 & Contradiction   & \begin{tabular}[c]{@{}c@{}}Entailment\\ (\textbf{right answer})\end{tabular}     \\
\hline
\hline
\begin{tabular}[c]{@{}l@{}}P: The patient was sent to the HD unit prior to coming to the floor for \textbf{workup} \textit{(A general}\\ \textit{medical examination to assess a persons health and fitness)} of fever.\\ H: The patient has an infection \end{tabular}                                                                 & \begin{tabular}[c]{@{}c@{}}Neutral\\ (\textbf{right answer})\end{tabular}    & Contradiction     \\
\hline
\begin{tabular}[c]{@{}l@{}}P: - \textbf{COPD} \textit{(chronic obstructive pulmonary disease)} - obesity - unspecified hypoxemia - \\CNS lymphoma c/b CVAs x3 (posterior circulation) and seizure d/o - history of SAH \\
while on coumadin - diastolic heart failure - coronary artery disease - atrial fibrillation -\\ hypertension - hyperlipidemia - severe OSA (did not tolerate CPAP in the past) - primary \\
hyperparathyroidism/25-vit D deficiency c/b nephrolithiasis - toxic multinodular \\
goiter with \textbf{subclinical} \textit{(Less than is needed for clinical reasons)} hyperthyroidism - \\ neovascular glaucoma c/b right eye blindness \\H: Patient has a history of malignancy \end{tabular}                                                                 & \begin{tabular}[c]{@{}c@{}}Neutral\\ (\textbf{right answer})\end{tabular}    & Entailment     \\
\hline
\hline
\end{tabular}
}
\caption[width=\textwidth]{Cases that model predicts differently after the supplement of paraphrases for rare biomedical words in MedNLI. ``P" for Premise and ``H" for Hypothesis. Words in \textbf{bold} are rare biomedical words and expressions in \textit{italic} inside the parentheses are the paraphrases of rare biomedical words.}
\label{case-study}
\end{table*}


\section{Conclusion}
Rare biomedical words are pervasive in biomedical texts, and understanding domain-specific rare words remains a tough challenge for pre-trained models.  In this work, we presented a simple yet effective method to help the pre-trained models grasp the semantics of rare biomedical words.

Enlightened by human reading behavior, we taught the pre-trained models to understand rare biomedical words by incorporating paraphrases of rare biomedical words. Our method can be regarded as a generic plug-in approach for prompt-based tuning without additional parameters. Experiments showed that our method could substantially improve the pre-trained models under few-shot settings.

\section*{Acknowledgements}

We thank the anonymous reviewers for their helpful and constructive comments and gratefully acknowledge the support of the National Key R\&D Program of China (2021ZD0113302) and the National Natural Science Foundation of China Youth Fund (62206079).

\bibliographystyle{acl_natbib}
\bibliography{ms}

\clearpage


\end{document}